\documentclass{article}

\PassOptionsToPackage{numbers}{natbib}
\usepackage[preprint]{neurips_data_2022}
\usepackage{graphicx}
\usepackage{amsmath}
\usepackage[hyphens]{url}
\usepackage[utf8]{inputenc} 
\usepackage[greek.polutonic, main=english]{babel}
\usepackage{makecell}
\usepackage[flushleft]{threeparttable}
\usepackage[T1]{fontenc}    
\usepackage{hyperref}       
\usepackage{booktabs}       
\usepackage{amsfonts}       
\usepackage{nicefrac}       
\usepackage{microtype}      
\usepackage{xcolor}         
\usepackage{authblk}
\usepackage{float}
\usepackage{cleveref}
\crefformat{footnote}{#2\footnotemark[#1]#3}
\begin{document}
\title{A Greek Parliament Proceedings Dataset for Computational
  Linguistics and Political Analysis}

\makeatletter
\renewcommand\AB@affilsepx{, \protect\Affilfont}
\makeatother
\author[1]{Konstantina Dritsa}
\author[2]{Kaiti Thoma}
\author[3]{John Pavlopoulos}
\author[4]{Panos Louridas}
\affil[1,2,3,4]{Athens University of Economics \& Business, Greece}
\affil[3]{	Stockholm University, Sweden}
\affil[1]{dritsakon@aueb.gr}
\affil[2]{aikelthoma@gmail.com}
\affil[3]{annis@aueb.gr}
\affil[4]{louridas@aueb.gr}

\maketitle

\begin{abstract}
  Large, diachronic datasets of political discourse are hard to come
  across, especially for resource-lean languages such as g In
  this paper, we introduce a curated dataset of the Greek Parliament
  Proceedings that extends chronologically from 1989 up to 2020. It
  consists of more than 1 million speeches with extensive metadata,
  extracted from 5,355 parliamentary record files. We explain how it
  was constructed and the challenges that we had to overcome. The
  dataset can be used for both computational linguistics and political
  analysis---ideally, combining the two. We present such an
  application, showing (i) how the dataset can be used to study the
  change of word usage through time, (ii) between significant
  historical events and political parties, (iii) by evaluating and
  employing algorithms for detecting semantic shifts.

\end{abstract}

\section{Introduction}
\label{section:introduction}
\vspace{-1em} The meanings of words change continuously through time,
reflecting the evolution of societies and ideas. For example the word
``gay'' originally meant ``joyful'' but gradually changed its usage to
refer to sexual orientation~\cite{kulkarni:2015}. Perhaps less
well-known, but probably more relevant to our subject, in 1850
rubbish-tip pickers were using the term ``soft-ware'' for material
that will decompose and ``hard-ware'' for the rest~\cite[p.\
380]{dickens:1850}. The availability of large corpora and advances in
computational semantics have formed fertile ground for the study of
semantic shifts. When these advances are applied to parliamentary
corpora, they can provide useful insights into language
change\footnote{``Language evolution'', ``lexical semantic change'',
  ``terminology evolution'', ``semantic change'', ``semantic shift'',
  and ``semantic drift'' are also all terms used for the same
  concept.}, different political views, and the effect of historical
events to the use of a language. 

In this paper, we present a curated diachronic Greek language dataset,
extracted from the proceedings of the Greek Parliament and spanning 31
years. It consists of more than 1 million speeches in chronological
debate order, with extensive metadata about the speakers, such as
gender, political affiliation, and political role. To our knowledge,
it is the only freely available dataset covering a comparable length
of time in the Greek language. Moreover, by its nature as a record of
the country's parliament, it is again to our knowledge the only
dataset that captures more than a quarter century of the recent Greek
political history. We demonstrate the value of the dataset by using it
to evaluate four state of the art word usage change detection
approaches and select the most appropriate among them to compute
word usage changes across time and among political parties.

The paper is organized as follows:
Section~\ref{section:related_work} summarizes the
approaches for word usage change detection.
Section~\ref{section:dataset_description} presents our dataset and
its construction process. In Section~\ref{section:methodology} we
evaluate four state of the art word usage change detection algorithms.
In Section~\ref{section:evaluation} we
examine how word usage changes reflect political events in Greece.
Section~\ref{section:conclusions} presents our conclusions and further
discussion. \vspace{-2mm}

\section{Related Work, Challenges, and the Parliamentary Dataset}
\label{section:related_work}
\vspace{-3mm} Researchers have attempted to capture diachronic
semantic shifts of words with the use of distributional semantics,
based on the distributional hypothesis~\cite{firth:1957}. According to
this hypothesis, each word is characterized by the company it keeps.
It follows that the change in the usage of word, that is, its semantic
shift, is defined by the change in the words co-occurring in its
context. Computationally, words are embedded in short dense vectors
according to their co-occurrence relationships and word usage change
can be measured by the distance between vectors that are calculated on
data of different time periods~\cite{azarbonyad:2017}. Approaches of
capturing diachronic semantic shifts can be divided into
projection-based and neighbor-based~\cite{gonen:2020,kutuzov:2018}.
The former have shown to be mostly suitable for detecting changes of
linguistic drift, more prominent in verbs, while the latter for
capturing cultural semantic shifts, encountered more frequently in the
nominal domain~\cite{kutuzov:2018,hamilton:2016:cultural}.

According to the projection-based approach, word vectors calculated on
different corpora are projected in a shared vector space and usage
change is computed with the cosine distance. However, vectors trained
on different corpora are not comparable by default, as word embedding
algorithms are inherently stochastic. Thus, many transformation
methods have emerged, with the most prominent being vector space
alignment~\cite{hamilton:2016:procrustes, mikolov2013, kulkarni:2015,
  kim:2014,peng:2017}. Hamilton et al.~\cite{hamilton:2016:procrustes}
(hereafter \textbf{``Orthogonal Procrustes''}) use orthogonal
Procrustes transformations to align diachronic models. Recent studies
are still building on this work: such is the case of incremental
update methods, where one trains a model on one corpus and then
updates it with data from the other corpus, while saving its state
every time~\cite{kim:2014,peng:2017}.. Carlo et
al.~\cite{dicarlo:2019} (hereafter \textbf{``Compass''}) build on the
assumption that it is the context of the word that changes over time,
but the meanings of the individual words in each different context
remains relatively stable. From that assumption they work with the
context embeddings learned by word2vec
models\cite{mikolov2013word2vec}, trained on atemporal target
embeddings that function as an alignment compass. 

The neighbor-based approach uses directly the different neighboring
words that reflect change. Gonen et al.~\cite{gonen:2020} (hereafter
\textbf{``NN''}) introduce intersection@$k$, i.e., the intersection of
the word's top-$k$ nearest neighbors from each corpus, to measure the
difference of neighboring words. In their work, they propose
that projection-based methods are more sensitive to proper nouns.

Hybrid approaches have emerged, combining properties of the
aforementioned categories. Hamilton et al.~\cite{hamilton:2016:cultural}
(hereafter \textbf{``Second-Order Similarity''})
collect the word union of the top-$k$ neighbors of a word~$w$ from two
different corpora. Then, they create a second-order embedding for each
corpus with the similarity between~$w$ and each neighbor. Intuitively,
usage change is estimated by the angle the word's neighborhood has to
cover when moving from one corpus to the other. In their work
they propose that cultural changes should be studied with
neighbor-based approaches.

Furthermore, the rise of contextual embeddings such as
BERT~\cite{devlin:2019} and ELMo~\cite{peters:2018} has enabled
important developments in the study of word usage change as they are
capable of generating a different vector representation for each
specific word usage. Contextual embeddings can be used in the context
of usage change detection by aggregating the information from the set
of token
embeddings~\cite{montariol:2021,liu:2021,martinc:2020a,kutuzov:2020,giulianelli:2020}.
However, related work shows that, for the time being, it is complex to
disambiguate between word senses, and there is a large disparity
between results on different
corpora~\cite{martinc:2020a,liu:2021,montariol:2021,laicher:2021}.
Finally, recent studies have emerged that ensemble multiple types of
word embeddings and distance metrics to experiment on improving
overall performance~\cite{kutuzov:2020,martinc:2020b}.

Different approaches can give different results, thus comparing them
is a challenge~\cite{shoemark:2019}. An additional challenge is the
stability of the approach used. An approach demonstrates stability
when slightly different runs on a dataset do not significantly affect
the results~\cite{gonen:2020}. Recent studies highlight the importance
of stability, as a high variation can be a sufficient reason to call
the whole method into question~\cite{antoniak:2018,burdick:2021}.
Researchers have identified a number of factors that affect stability,
including properties of the underlying algorithms used to construct
the
embeddings~\cite{gonen:2020,wendlandt2018,antoniak:2018,levy2015,hellrich2016}.
Subsequent runs of word embedding algorithms on the same data will not
necessarily produce the same results, due to the stochastic nature of
the approaches. Gonen et al.~\cite{gonen:2020} use intersection@$k$,
mentioned above, to gauge the stability between the predictions of two
different runs of the same algorithm. We adopted this metric,
in order to select a stable usage change algorithm for our
study.

Existing studies on language change use corpora of high resource
languages such as English, German, French, Spanish, and Chinese,
spanning centuries~\cite{sagi:2011,helsinkicorpus:1991,davies:2015} or
decades, comprising tweets and product reviews~\cite{kulkarni:2015},
digital books~\cite{michel:2011} and news
articles~\cite{nytcorpus:2008,enggigaword:2011}. In English, a work
similar to ours is that of Azarbonyad et al.~\cite{azarbonyad:2017},
in which they study the semantic shift of words in the British House
of Commons. Also in English, Gentzkow et al.~\cite{uscongress:2018}
curated a dataset of the US Congress speeches from 1873--2017, with
extensive metadata on speeches and speakers.

In this work, we present an extensive dataset that can be used for the
study of language change in the context of the Greek Parliament. To
the best of our knowledge, there are no existing computational
studies on language change in modern Greek.
We show the value of the dataset by utilizing
it to comparing four state-of-the-art approaches of language change
detection, namely Orthogonal
Procrustes~\cite{hamilton:2016:procrustes},
Compass~\cite{dicarlo:2019}, NN~\cite{gonen:2020} and Second-Order
Similarity~\cite{hamilton:2016:cultural}. The selection of the
approaches for language change detection aims to be representative of
different established methodologies proposed in the related work and
does not constitute a complete benchmark evaluation on language change
detection methods. Furthermore, following the challenges identified
above, we evaluate the stability of the approaches using the
intersection@$k$ measure. We also qualitatively evaluate their results
on the change of word usage between the decades 1990--1999 and
2010--2019. Finally, as the dataset is a mirror of political history,
we use it to detect word usage changes between different time periods,
before and after important historical events, as well as among
different political entities. \vspace{-1mm}

\section{Dataset Description and Construction}
\label{section:dataset_description}
\vspace{-1mm}

\subsection{Contents}
\label{section:contents}
\vspace{-1mm}
The dataset\footnote{\url{https://zenodo.org/record/7005201}}
includes 1,280,918 speeches of parliament members in
chronological debate order, exported from 5,355 parliamentary sitting record
files, with a total volume of 2.12 GB. The speeches extend
chronologically from July 1989 up to July 2020.
Table~\ref{table:dataset-fields} shows the contents of the dataset.
\vspace{-1mm}
\begin{table}[!htbp]
	\scalebox{0.95}{%
  \begin{tabular}{|>{\bf}p{0.25\linewidth} p{0.7\linewidth}|}
  \hline  
  \textbf{member\_name} & the name of the person speaking\\ \hline
  sitting\_date & the date the sitting took place\\ \hline
  parliamentary\_period & the name and/or number of the
  parliamentary period that the speech took place in. A parliamentary
  period is defined as the time span between one general election and
  the next. A parliamentary period includes multiple parliamentary
  sessions.\\ \hline
  parliamentary\_session & the name and/or number of the
  parliamentary session when the speech took place. A session is
  a time span of usually 10 months within a parliamentary
  period during which the parliament can convene and function as
  stipulated by the constitution.  A
  parliamentary session includes multiple parliamentary sittings.\\ \hline
  parliamentary\_sitting &  the name and/or number of the
  parliamentary sitting that the speech took place in. A sitting is
  a meeting of parliament members.\\ \hline
  political\_party & the political party of the speaker\\ \hline
  government & the government in power when the speech took place\\ \hline
  member\_region & the electoral district the speaker belonged to\\ \hline
  roles & information about the speaker's parliamentary and/or
  government roles\\ \hline
  member\_gender & the gender of the speaker\\ \hline
  speech & the speech delivered during the parliamentary sitting\\
  \hline 
\end{tabular}
}
\caption{Contents of the Parliament Proceedings Dataset}  
\vspace{-3mm}
\label{table:dataset-fields}
\end{table}

\begin{figure}[t]
  \centering
  \centerline{\includegraphics[scale=0.4]{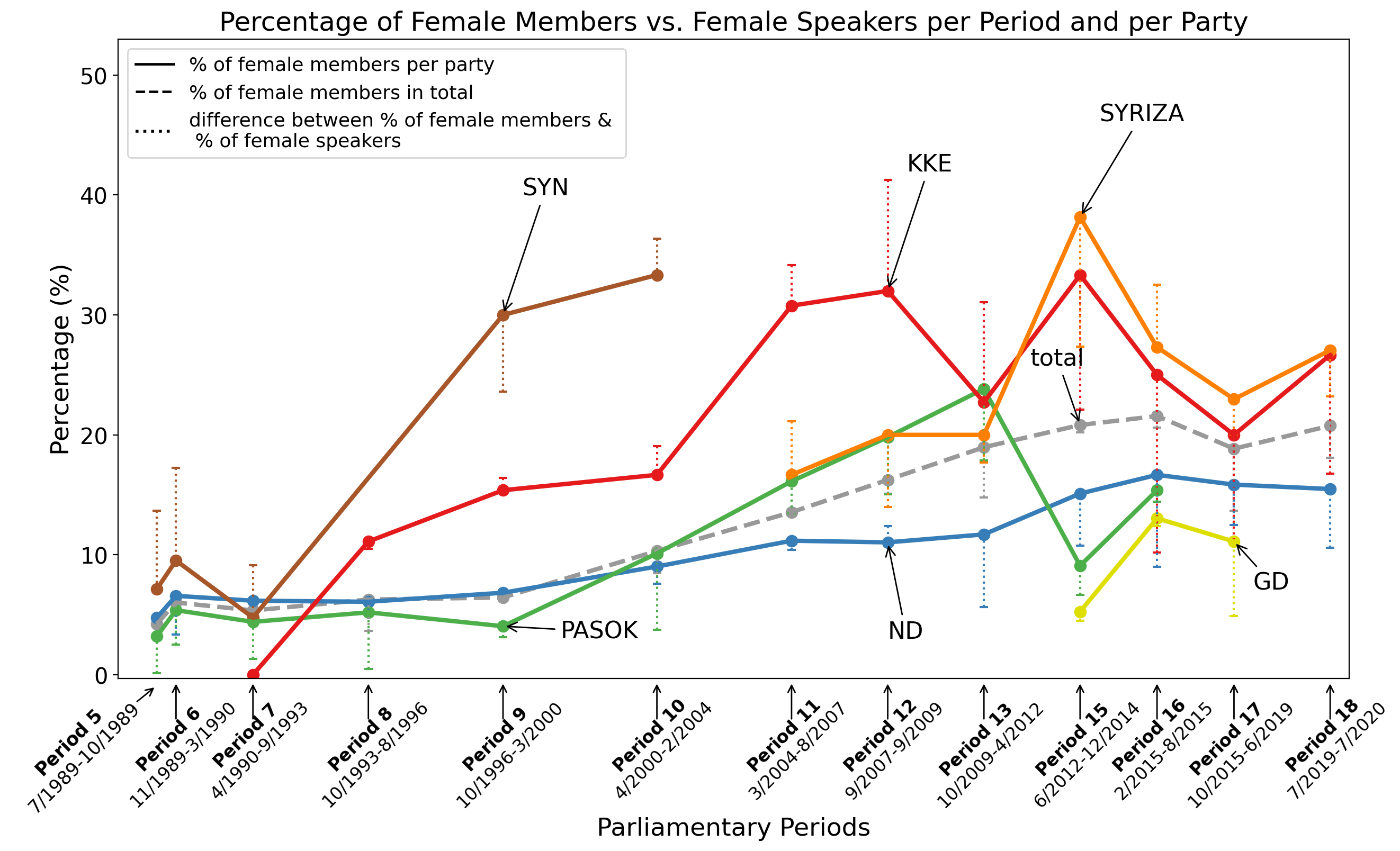}}
  \caption{Percentage of female members and corresponding speech
    activity per period.}
	\vspace{-4mm}
  \label{fig:gender_per_period}
\end{figure}

Delving deeper into our dataset, Fig.~\ref{fig:gender_per_period}
depicts the percentage of female members in the Greek Parliament and
the percentage of characters of speech delivered by female
individuals, per political party and per parliamentary period. The
difference between the membership percentage and the speech percentage
is highlighted with dotted vertical lines. For reasons of readability,
we depict political parties that have played an important role in
recent political history. These are New Democracy (center-right,
hereafter ND), the Panhellenic Socialist Movement (center-left,
hereafter PASOK), the Coalition of the Radical Left---Progressive
Alliance (left, hereafter SYRIZA), the Communist Party of Greece
(communist, hereafter KKE), the Coalition of the Left, of Movements
and Ecology (left, hereafter SYN) and Golden Dawn (extreme right,
nationalist, nazi-fascist, hereafter GD). We exclude period 14, which
lasted two days and was a transitional government.

Concerning gender representation in the parliament, the total
percentage of female members (dashed line) increases over time.
Left-wing political parties like SYN, KKE and SYRIZA achieve higher
percentages of female members and remain above the total percentage of
female members for almost all periods. The percentage of center-left
PASOK presents great fluctuation over the years. On the other hand,
the percentage of the center-right ND remains below the total average
percentage. Lastly, the far-right GD has the lowest percentage of
female members of parliament, compared to the selected political
parties. Regarding the participation of females in parliamentary
debates, only the left-wing SYN and KKE achieve percentages higher
than that of female membership for most periods. Overall, none of the
examined parties has a percentage of female members equal to or
greater than 50\% at any point in time. After investigating the rest
of the parties, we found that only two left-wing parties have achieved
percentages greater than 50\% for female members, namely Alternative
Ecologists (greens, left, 100\% for periods 6 \& 7) and MeRA25 (left,
55\% for periods 16 \& 18). \vspace{-1mm}
\subsection{Record Collection}
\vspace{-1mm} Due to the absence of an API, we crawled the catalogue
of parliament records from 1989 up to 2020, available from the
official Greek Parliament
website\footnote{\url{https://www.hellenicparliament.gr/Praktika/Synedriaseis-Olomeleias}.
  The proceedings for 1995 are not publicly available.}.
Files were in doc, docx, text, PDF or HTML format. We converted all
to text using Apache
Tika\footnote{\url{https://tika.apache.org/download.html},
  tika-app-1.20.jar}. 

Each record of a parliamentary sitting begins with some introductory
information, followed by the debate that took place. Typically, each
speaker's full name is written in capital letters at the beginning of
a new line, followed by a colon and the corresponding speech. The name
is occasionally accompanied by a parenthesis with information about
the person, such as their political party or governmental role.
Unfortunately, the records contain material that fails to follow this
format. So, to extract speeches from the parliamentary records it was
necessary to create, in a preliminary step, a number of auxiliary
datasets as described below.
\vspace{-1mm}
\subsection{Support Datasets}
\vspace{-1mm}
\paragraph{Female \& Male Names} 
We crawled the entries of the Wiktionary Greek names
category\footnote{\url{https://en.wiktionary.org/wiki/Category:Greek_names}}
and created a support dataset of modern Greek female and male names
and surnames and their grammatical cases, filling missing entries
using the rules of grammar. 
\vspace{-1em} 
\paragraph{Elected Members of Parliament}
The Greek Parliament website provides a
list\footnote{\url{https://www.hellenicparliament.gr/Vouleftes/Diatelesantes-Vouleftes-Apo-Ti-Metapolitefsi-Os-Simera/}}
of all the elected members of parliament since the fall of the
military junta in Greece, in 1974. For each member, we extracted the
exact date range of their activity in each political party during each
parliamentary period. We added the gender of each member, based on the
gender of their name from the ``Female \& Male Names'' dataset.
\vspace{-1em}

\paragraph{Government Members}
As government members we refer to individuals in ministerial or other
government posts, regardless of whether they were elected in the
parliament. This information is available in the website of the
Secretariat General for Legal and Parliamentary
Affairs\footnote{\label{gslegal}\url{https://gslegal.gov.gr}}. Names
and surnames are given in the genitive case and cannot be matched
directly to parliamentary records, where names are given in the
nominative case. To resolve this, we used the ``Female \& Male Names''
dataset to convert the collected genitive cases to nominative and
deduce the gender.\vspace{-1em}

\paragraph{Governments} We automatically collected from the website of the
Secretariat General for Legal and Parliamentary
Affairs\cref{gslegal} a support dataset with the
names of governments since 1989, their start and end dates, and a URL
that points to the respective official government web page of each
government. \vspace{-1em}

\paragraph{Additional Political Posts}
We manually collected from Wikipedia additional government and
political posts that were not included in the previous resources:
service information of the Chairmen, Speakers and Deputy Speakers of
the Parliament, party leaders, and opposition leaders.\vspace{-1em}

\paragraph{Merged Support Dataset} 
We merged the above datasets producing an integrated file. Each row of
the final file includes the full name of the individual, the start and
end date of their term of office, the political party and electoral
district they belonged to, their gender, the parliamentary and/or
government positions that they held along with start and end dates,
and the name of the government that was in power during their term of
office.
\vspace{-1mm}
\subsection{Speech Extraction}
\label{section:speech_extraction}
\vspace{-1mm}

\paragraph{Speaker Detection}
To identify each new speech, we had to identify a valid candidate
speaker. As mentioned, in many cases the text did not follow the
expected format. For example,
some new speeches would not start at the beginning of a new line or
there would be missing closing brackets in the speaker's reference.
We created a comprehensive list of regular expressions in order
to capture possible debate formats. 
\vspace{-1mm}
\paragraph{Entity Resolution}
After the detection of a candidate speaker, we matched the extracted
speaker to our list of individuals with the use of the
Jaro-Winkler~\cite{winkler:1990} string similarity metric. However,
although not as problematic as characters in a Russian novel, there
exist many different name variants in the records. Some speakers were
referenced with their nicknames. For people with more than one
names/surnames, some of them where missing and the order of the
first/last names was not always the same. To resolve this string
comparison task, we created all possible variants of an official name,
alternating the order of the words that make up that name and
replacing or combining the name with its variants. Due to
misspellings, we accepted matches with similarity $\ge 0.95$. For
matching we used yet another dataset of 475 names and nicknames, which
cannot be shared due to licensing reasons.
\vspace{-1mm}

\subsection{Preprocessing}
\vspace{-1mm}
We replaced all references to political parties with the symbol ``@''
followed by an abbreviation of the party name.
We removed accents, strings with length less than 2 characters, all
punctuation except full stops, and replaced stopwords with ``@sw''.

\begin{figure}[t]
  \centering
  \centerline{\includegraphics[scale=0.4]{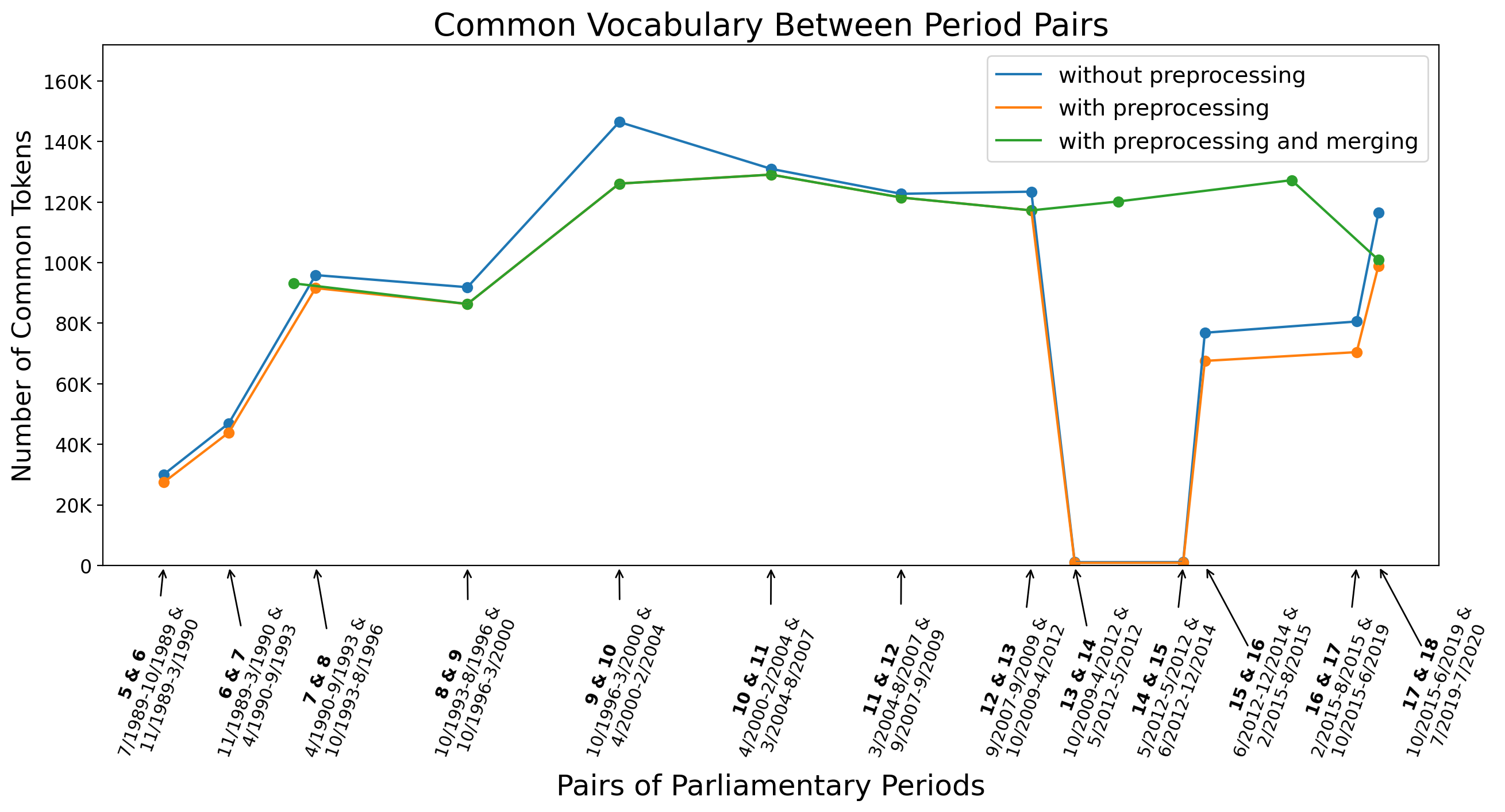}}
  \vspace{-2mm}
  \caption{Common vocabulary between consecutive pairs
    of periods before preprocessing, after preprocessing
    and upon preprocessing and merging small periods with the next one. }
  \label{fig:common_vocab}
\end{figure}

The volume of data per parliamentary period varies, as does the shared
vocabulary between consecutive periods. This is key to our
investigation, as commonalities between the vocabularies across time
are necessary to detect usage change. Fig.~\ref{fig:common_vocab}
shows the common vocabulary in terms of tokens between pairs of
consecutive periods. Periods 5, 6, 14 and 16 are
transitional and span between 1 to 7 months, resulting in low
vocabulary overlap with other periods.
In these cases of small shared vocabulary, important semantic shifts
are usually artifacts of the lack of data. To avoid biased
conclusions, we merged the smaller periods with their
following large period, these being period 5 and 6 with
period 7, period 14 with 15 and period 16 with 17.
\begin{table}[t]
  \caption{Average metrics for each parliamentary period before
    preprocessing, after preprocessing, and after merging the small
    periods with their consecutive large periods.}
  \label{tab:stats_1} 
  \centering
  \resizebox{\textwidth}{!}{%
    \begin{tabular}{cccccc}
      \hline
      &
        \textbf{\begin{tabular}[c]{@{}c@{}}Avg. characters\end{tabular}}
      &
        \textbf{\begin{tabular}[c]{@{}c@{}}Avg. tokens\end{tabular}}
      & \textbf{\begin{tabular}[c]{@{}c@{}}Avg. unique tokens\end{tabular}}
      &
        \textbf{\begin{tabular}[c]{@{}c@{}}Avg. sentences\end{tabular}}
      & \textbf{\begin{tabular}[c]{@{}c@{}}Avg.
                  unique sentences\end{tabular}} \\ \hline
      \textbf{No preprocessing} & 82.4M & 14.18M & 208.12K & 570.11K
      & 516.99K \\ \hline
      \textbf{With preprocessing} & 78.22M & 19.57M & 216.79K
      & 416.46K & 386.83K \\ \hline
      \textbf{\begin{tabular}[c]{@{}c@{}}With preprocessing\\
                \& merged periods\end{tabular}}
      & 109.51M & 27.4M & 287.19K & 583.04K & 541.04K \\ \hline
    \end{tabular}
  }
\vspace{-5mm}
\end{table}
Table~\ref{tab:stats_1} shows descriptive statistics of the dataset in
three different steps, before and after preprocessing, and upon
preprocessing and merging the smaller periods with their following
large periods. Preprocessing leads to a decrease of the average number
of characters and sentences but increases the average tokens and
unique tokens. Merging the periods increases all numbers.
\vspace{-2mm}

\section{Quantitative and Qualitative Evaluation}
\label{section:methodology}
\vspace{-1mm}
\subsection{Quantitative Evaluation: Stability}
\label{subsection:stability}
\vspace{-2mm} We compared stability between Orthogonal
Proctustes~\cite{hamilton:2016:procrustes},
Compass~\cite{dicarlo:2019}, NN~\cite{gonen:2020} and Second-Order
Similarity~\cite{hamilton:2016:cultural}, as well as a variation of
the Compass method in which we introduced the frequency cut-offs of
the NN approach~\cite{gonen:2020}. Specifically, we removed from the
vocabulary of each model the 200 most frequent words and words that
appear less than 200 times.
In our case, the frequency distribution for each
corpus is long-tailed, with only $\sim$5\% of the vocabulary of each
decade having 200 or more occurrences.
Our aim is to investigate whether the removal of these words
might increase the stability of the results.

We applied the comparison on the task of word usage change detection
between 1990--1999 and 2010--2019. We used intersection@$k$, proposed
by Gonen et al.~\cite{gonen:2020}, which measures the percentage of
shared words in the $k$ most changed words for a number of restarts,
each time changing the random seed. For each approach (e.g., Compass)
and between the two time periods, we measured word usage change and
detected the most changed words. By repeating the measurement with
different random seeds, then, we computed the common changed words
across repetitions. Specifically, we ran each usage change approach 10
times and collected the top-$k$ most changed words, where
$k \in [10, 20, 50, 100, 200, 500, 1000]$. Then, for each of the
$\binom{10}{2} = 45$ pairs of different runs and for each of the
values of $k$, we measured the percentage of shared words in the
most-changed-words predictions of each approach. A value of zero
between a pair of runs means there are no shared words in their
predictions, indicating high variability, while a value of one
indicates high stability.

\begin{figure}[t]
  \centering
  \centerline{\includegraphics[scale=0.35]{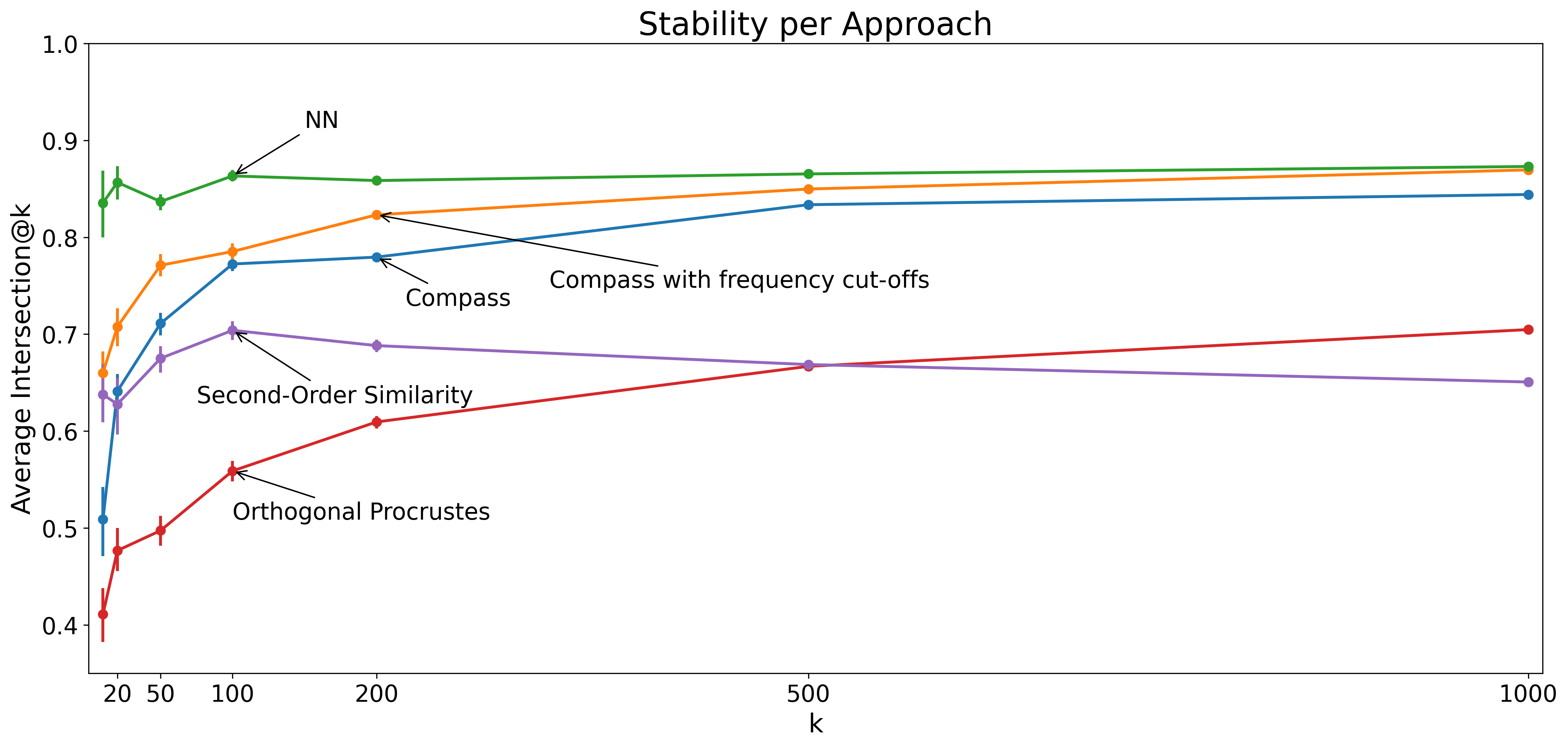}}
  \vspace{-2mm}
  \caption{The average intersection@$k$ for 45 pairs of different runs
    and for different values of $k$. }
  \label{fig:stability_comparison}
  \vspace{-1mm}
\end{figure}
\noindent
Fig.~\ref{fig:stability_comparison} shows the average intersection@$k$
between the 45 pairs of different runs for each approach, with 95\%
confidence intervals calculated by the bootstrap method. The NN method
exhibits the greatest stability even for very small values of $k$.
This means that for all values of $k$ and regardless of the random
seed used, the changed words this approach detected were mostly the
same. Compass follows closely, while the Compass variation yields
better stability results, similar to that of NN. The Orthogonal
Procrustes and the Second-Order Similarity approach gave worse
stability results, with the latter even decreasing for $k > 100$.

\vspace{-2mm}
\subsection{Qualitative Evaluation: Top Changed Words between 1990-1999 \&
  2010-2019}
\vspace{-1mm}

We qualitatively evaluated the top 100 most changed words between the
decades 1990--1999 and 2010--2019, as detected by each approach.
We introduce a frequency threshold of 50 occurrences in at least one
of the two decades, for the approaches that do not already have
frequency thresholds.
\begin{table}[]
  \caption{A representative selection of words from the top 100 most changed
  	words per approach between the 1990s and 2010s.}
  \label{table:topchanged_allmethods}
  \centering
  \scalebox{0.9}{%
    \begin{tabular}{|c|c|}
      \hline \multicolumn{1}{|l|}{}  & \textbf{Selection of top
                                       changed words} \\ \hline
      \textbf{Compass}  & \makecell{psi, haircut, normality, vatopedi,
                          cook}
      \\ \hline
      \textbf{\makecell{Compass variation}} & \makecell{agenda,
                                              inputs,
                                              european economic community,
                                              drachma, green} \\ \hline
		\textbf{NN}  & \makecell{simple, deny, called, people,
                               interested} \\ \hline
      \textbf{\makecell{Orthogonal Procrustes}} & \makecell{red,
                                                  clarity,
                                                  capital, Prespa,
                                                  migratory} \\ \hline
      \textbf{\makecell{Second-Order Similarity}} & \makecell{give,
                                                    phthiotis,
                                                    laconia,
                                                    arcadia, critical}
      \\ \hline
    \end{tabular}
  }
\vspace{-3mm}
\end{table}
Table~\ref{table:topchanged_allmethods}, shows a representative
selection of results for each approach. 

Compass detected words that have meaningful change connected with
Greek historical events. The words ``psi'', ``haircut'',
``normality'', ``agenda'', ``drachma''
are largely related to the Greek financial crisis of the 2010s.
``PSI'' stands for ``Private-Sector Involvement'', meaning that
private investors had to accept a write off on the face value of Greek
government bonds they were holding. A ``haircut'' is a cut to existing
debt. ``Vatopedi'' referred to an economic scandal involving an
homonymous monastery.
``Cook'' was connected to the bankruptcy of the British travel firm
``Thomas Cook'', possibly affecting the Greek tourist industry.

Orthogonal Procrustes also detected word usage changes that make sense
in the historical context. ``Red'' was used for the so-called ``red
loans'', non-performing loans that emerged during the crisis.
``Clarity'' became connected to the online platform
\url{diavgeia.gov.gr}, where government spending is publicly published
to improve transparency. ``Capital'' in the 2010s referred to the
capital controls applied in the Greek banks. ``Prespa'', name of a
lake, was used in the 2010s to describe the Prespa agreement between
Greece and the Republic of North Macedonia.
Change in the word ``migratory'' reflects the increased arrivals of
refugees by sea in the 2010s, mainly due to the Syrian civil war.

NN and Second-Order Similarity approaches provided less explainable
results. The top-100 list of NN included mainly verbs and no proper
nouns. The closest neighbors of the verbs consisted mainly of
grammatical persons, tenses and synonyms. The Second-Order Similarity
results included almost only geographical regions, numbers and names
of months.
\vspace{-1mm}
\vspace{-1mm}

\section{Changes in Word Usage and Political History}
\label{section:evaluation}
\vspace{-1mm}

As Compass presented the best performance combination in both stability and
detection of meaningful change, we used it to investigate changes in
word usage and events in Greece's recent political history.
\vspace{-5mm}
\subsection{Top Changed Words before and after the Greek Economic
  Crisis} 
\vspace{-2mm}

Greece faced a threat of sovereign default in 2007--2008, leading to
a massive recession.
In the
following years, Greek governments adopted austerity measures in a
series of adjustment programs agreed with Eurozone countries and the
International Monetary Fund (IMF). We detect word usage changes
between the decades before ($t1$: 1997--2007) and during ($t2$:
2008--2018) the crisis.

\begin{table}[!ht]
  \caption{Words with notable usage change before ($t1$) and during
    ($t2$) the Greek economic crisis. 
  }
  \label{table:top_changed_crisis}
  \centering
  \resizebox{\textwidth}{!}{%
    \begin{tabular}{|c|c|c|c|}
      \hline
      \textbf{Word} & \textbf{Similarity}
      & \textbf{Neighbors @ $t1$}
      & \textbf{Neighbors @ $t2$}\\ \hline
      haircut & -$0.06$ & \makecell{gypsy, sixteen-year-old, excellent, 
                        empirical}
      & \makecell{psi, repurchase, haircut, reduction} \\ \hline
      psi & $0.01$ & \makecell{boilers, rented, fainted, humanization}
      & \makecell{haircut, repurchase, 	bonds, sector} \\ \hline
      golden & $0.01$ & \makecell{feed, renegotiate, people, pretend}
      & \makecell{boys, platinum, chicago, hall} \\ \hline
      story & $0.01$ & \makecell{tortures, old-fashioned, nail,
                       tobaccoworker}
      & \makecell{success, true, fairy tale, myth} \\ \hline
      success & $0.04$ & \makecell{dried, liberated, interbank,
                         emerging}
      & \makecell{story, myth, make up, fairy tale} \\ \hline
      brain & $0.05$ & \makecell{distinguished, overpay, collected,
                       dermatological}
      & \makecell{drain, gain, circulation, migration} \\ \hline
      cutter & $0.06$ & \makecell{tweaked, fighting, salvage,
                        rescuing}
      & \makecell{automatic, mechanism, account, infamous} \\ \hline
      systemic & $0.06$ & \makecell{autumn, short-term therapy,
                          shape, segmented}
      & \makecell{corrupted, media, unchecked, regime} \\ \hline
      imf & $0.06$ & \makecell{hall, bleed, multivariate, superset}
      & \makecell{sch\"{a}uble, troika, monetary, european commission} \\ \hline
      \makecell{counter-\\measures} & $0.10$ & \makecell{resin, legal
                                              policies, fainted,
                                              peaks}
      & \makecell{burdensome, painful, anti-popular, recessionary} \\ \hline
    \end{tabular}
  }
\end{table}

Table~\ref{table:top_changed_crisis} presents a selection of words
with notable word usage change, along with their closest neighbors, at
each of the two decades; as usage change is measured with the cosine
similarity, low values represent significant change. ``Haircut'' did
not initially refer to ``a debt haircut''; PSI was a unit of pressure
in $t1$, before referring to private sector involvement in Greek bonds
write off. ``Cutter'' was used to describe measures for economic
stability, such as unemployment allowances. ``Golden'' in $t2$ became
part of the phrase ``golden boys'', referring to people working in
senior management positions with high incomes and provocative
lifestyles, whose administrative decisions usually burdened their
companies. It was also used in references that liken the crisis
policies with those of the Chicago Boys, the Chilean economists of the
Pinochet rule educated at the University of Chicago. ``Success'' and
``story'' were regularly employed together to ironically describe
government promises of economic prosperity. ``Brain'', appearing in
``brain drain'', referred to the migration of highly skilled people to
other countries in search for better living conditions. The word
``systemic'' was commonly used to negatively characterize mainstream
media that, while heavily indebted themselves, were supporting
government policies. ``IMF'' was used during the crisis in the context
of the strict financial reforms it required from the Greek government.
``Countermeasures'' referred to government's compensatory measures
against the economic austerity. \vspace{-2mm}

\begin{figure}[t]
  \centering
  \centerline{\includegraphics[scale=0.35]{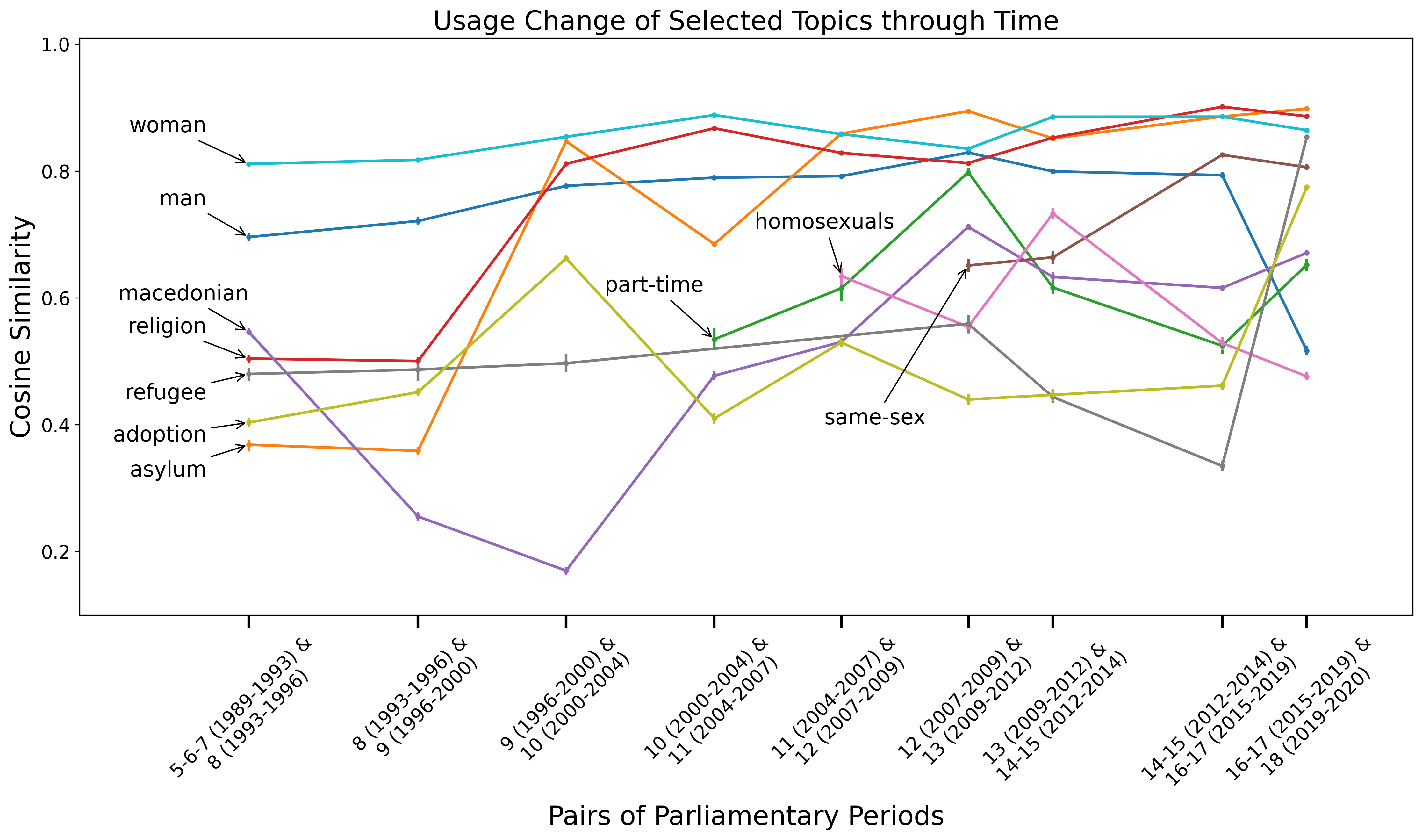}}
    \vspace{-2mm}
  \caption{Usage change of 10 political topics between pairs
    of parliamentary periods. Low cosine similarity denotes high
    usage change.}
  \label{fig:vouliwatch_periods}
  \vspace{-2mm}
\end{figure}
\noindent
\subsection{Usage Change of Popular Topics}
\label{rq2}
\vspace{-2mm} We estimate the usage change of selected topics that
were debated across periods. For the selection of topics, we consulted
the website of Vouliwatch\footnote{\url{https://vouliwatch.gr}}, a
non-partisan parliamentary monitoring organization that provides an
extensive comparison of party positions on 69 topics of significant
political interest. We extended this list with 22 additional topics,
selected for their popularity\footnote{The initial and extended lists
  of topics are available in the supplementary material of the
  paper.}. We repeated the usage change computations with 50 different
random seeds and calculated 95\% confidence intervals with the
bootstrap method.

Fig.~\ref{fig:vouliwatch_periods} shows a subset of topic
embeddings that exhibit notable decrease in semantic similarity
($\le 0.65$) over at least one pair of consecutive periods.
The word ``macedonian'' changes around 2000 from referring to a
business consortium named ``Macedonian metro'', to portraying the
turbulence around the naming dispute between Greece and the Republic
of North Macedonia. The similarity drop between periods
14--15 and 16--17 corresponds to extensive debates that took place in
the parliament nearing the dispute resolution in 2019. The word
``refugee'' changed from referring to cheap labor to reflecting the
increased number of persons crossing the borders to seek asylum in the
EU. The notable drop around 2015 is in agreement with external data
recording an increased volume of refugee arrivals at the
time~\footnote{\url{https://data.unhcr.org/en/situations/mediterranean/location/5179}}.
During the economic crisis, from 2008 onwards, ``part-time'' changed
usage as it became a common employment practice to reduce salary
expenses. The terms ``homosexuals'' and ``same-sex'' emerge around
2007, reflecting an increasing social awareness. In the following
years, the terms undergo important usage change as they approach the
words ``marriage'', ``cohabitation agreement'' and ``adoption''. The
word ``man'' changes context in 2019 from describing a male of the
typical Greek family model or a criminal to referring to a policeman,
associated with arbitrary police behavior and brutality.
The word ``woman'' does not display notable usage change but we
include it for comparison with the word ``man''. ``Woman'' is
constantly correlated with the words ``mother'', ``child'',
``spouse'', ``family'', exhibiting a context limited to traditional
family relations. \vspace{-1mm}
\subsection{Usage Change of Political Party Name Embeddings}
\label{section:rq3_name_embeddings}
\vspace{-2mm}

We gauge the usage change between parliamentary periods
of the names of political parties that have played an important
role in recent political history, introduced in
Section~\ref{section:contents}.

As mentioned in Section~\ref{section:speech_extraction}, we replaced
all political party references with the symbol ``@" followed by an
abbreviation of the party name, using regular expressions that capture
grammatical cases and variations. We trained Compass between
consecutive pairs of parliamentary periods and computed the cosine
similarity between the vectors of political party names. We repeated the
computations with 50 different random seeds and calculated 95\%
confidence intervals with the bootstrap method.
\begin{figure}[!htbp]
	\centering
	\centerline{\includegraphics[scale=0.35]{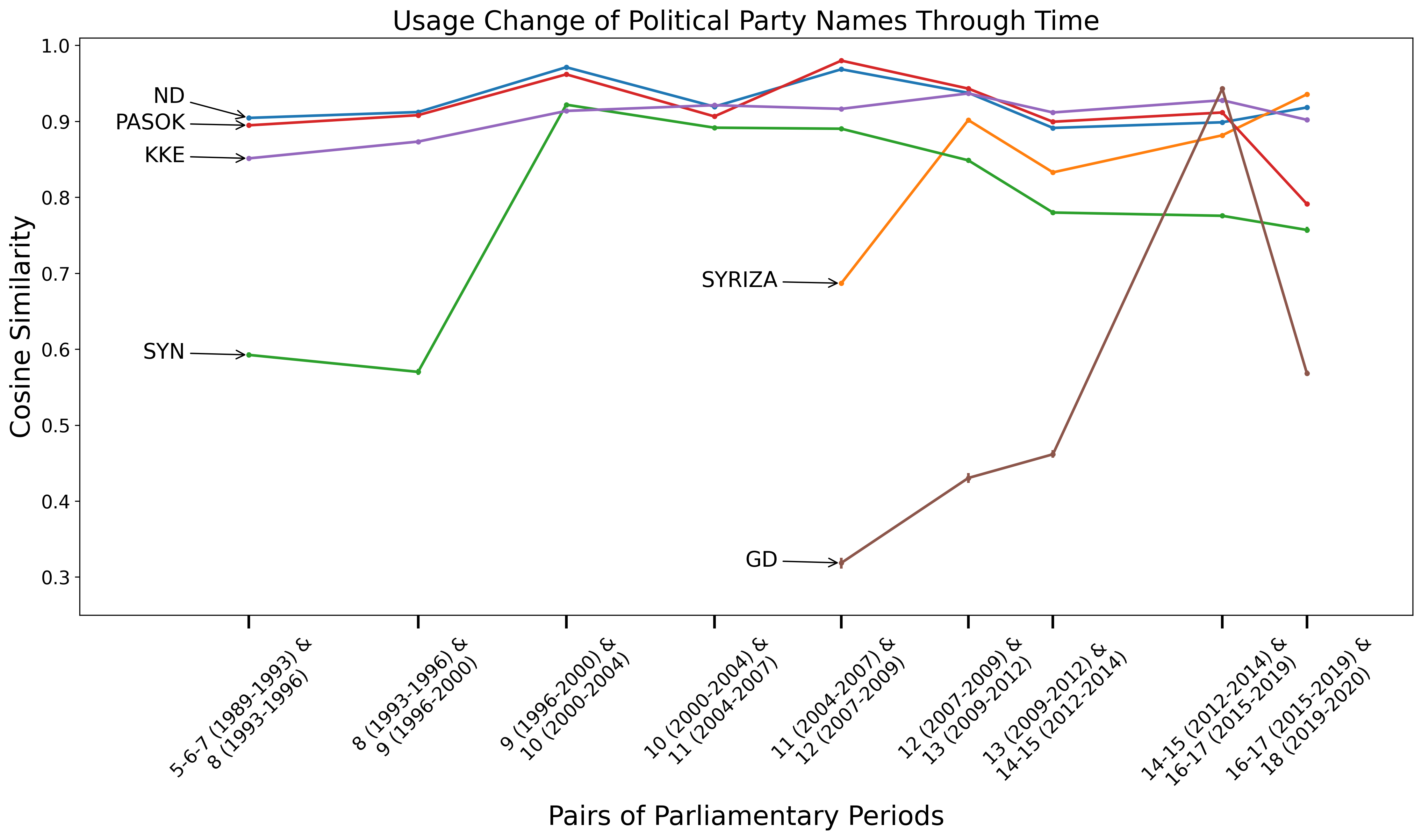}}
	\vspace{-2mm}
	\caption{Usage change of political party embeddings between pairs of consecutive periods.}
	\vspace{-1mm}
	\label{fig:party_embeddings}
\end{figure}
Fig.~\ref{fig:party_embeddings} presents the results. References to
political parties in the records through time do not reflect their
actual life-cycle. For example, although SYN was dissolved in 2013,
references to it persist in the following years.
ND, PASOK, and KKE show high similarity scores between all pairs of
consecutive parliamentary periods, reflecting a stable political
position. We locate the period pairs for which each party
embedding shows the lowest semantic similarity and study
its neighbors for each period to shed more detail
to their usage change.
During 2012--2014, ND appears closer to the words
``opposition'' and PASOK, as it was the official opposition party of
the government of PASOK. It is also close to the word ``Karamanlis'',
the name of the party leader at the time. During 2015--2019, ND comes
closer to the words ``coalition government'', PASOK and ``DIMAR'', an
abbreviation of the Democratic Left party, a minor left-wing political
party not shown in Fig.~\ref{fig:party_embeddings}. This change in
usage is consistent with political events of the period, when ND
formed a coalition government with PASOK and DIMAR. The usage change
of PASOK between the periods 2015--2019 and 2019--2020 coincides with
the incorporation of PASOK as the basic component of a new political
party, KINAL. The lower cosine similarity for KKE between 1989--1993
and 1993--1996 reflects the multiple coalitions and divisions it went
through at these periods (also reflecting effects in global history),
after which it remained in a stable state. For SYN, the usage change
between the periods 1993--1996 and 1996--2000 coincides with a crisis
brought by a failure to enter parliament, the resignation of the party
leader, and the election of new leadership. During 2004--2007, SYRIZA
is highly correlated with the name ``Alavanos'', the then party
leader, as well as SYN, the largest component of the alliance that
constituted SYRIZA. In 2007, Alavanos was succeeded by Tsipras. In the
following years, SYRIZA evolved from a loosely-knit coalition to the
largest party in parliament in 2015, leading a government coalition
with a minor partner (Independent Greeks, ANEL) in 2015--2019. That is
mirrored by highest cosine similarity, perhaps echoing a consistent
anti-austerity and anti-neoliberal message.
GD rose to prominence during the financial crisis. During its period
in the sun, GD was close to words like ``brutal'', ``beatings'',
``anarchist'', ``marches'', ``episodes'' and ``abusive'', reflecting
the criminal acts and attacks that perpetuated by supporters, members,
and high-ranking cadres of GD\@. Support for GD nosedived and did not
reach the 3\% threshold required to enter parliament in~2019.
\vspace{-1mm}
\noindent

\section{Conclusions and Future Work}
\label{section:conclusions}
\vspace{-3mm}
Large datasets of resource-lean languages on specific domains are hard
to find. In this work, we present a dataset of the Greek Parliament
proceedings spanning 31 years and consisting of more than 1 million
speeches, tagged with extensive metadata, such as speaker name,
gender and political role. We apply stable semantic
shift detection algorithms and detect notable word usage changes
connected with historical events such as the Greek economic crisis as
well as changes in the usage of political party names, connected with
internal organizational changes or election periods.

Our dataset has a specific provenance, parliamentary recordings, and
is not necessarily representative of language use and evolution in
general. Yet, it can be useful in various applications of
computational linguistics and political science, e.g., studies that
examine whether word usage change behaves differently in different
languages or contexts. Its extensive metadata can facilitate
fine-grained semantic change studies, such as to evaluate whether a
new parliament member gradually adjusts their speech to the style of
the majority of speakers. It can be used for monitoring and tracking
events and controversial topics over
time~\cite{kutuzov:2017,huang:2017,vaca:2014,emamgholizadeh:2020,dori:2016}
as well as rapid discourse changes during crisis
events~\cite{stewart:2017}, or for classification of political
texts~\cite{kusner:2015}. Other applications can include political
perspective detection~\cite{zhang:2022} and viewpoint
analysis~\cite{azarbonyad:2017} between parties, roles or genders and
cross-perspective opinion mining~\cite{fang:2012, ren:2016}. The
dataset can be combined with datasets of tweets and public statements
of parliament members, for modeling voting behavior and improving the
tasks of roll call vote and entity stance
prediction~\cite{mou:2021,feng:2022,yang:2020}.

\paragraph{Supplementary Material}
Download links to the original proceeding records, source code files and
implementation details are included in
the supplementary material that accompanies this paper.
The repository for this work is \url{https://github.com/Dritsa-Konstantina/greparl}.
\vspace{-2mm}
\paragraph{Acknowledgments}
This work was supported by the European Union’s Horizon
2020 research and innovation program ``FASTEN'' under
grant agreement No 825328 and the non-profit data
journalism organization iMEdD.org.


\bibliography{main.bib}
\bibliographystyle{plainnat}

\end{document}